\PassOptionsToPackage{unicode}{hyperref}
\PassOptionsToPackage{hyphens}{url}
\documentclass[
  11pt,
]{article}
\usepackage{amsmath,amssymb}
\usepackage{iftex}
\ifPDFTeX
  \usepackage[T1]{fontenc}
  \usepackage[utf8]{inputenc}
  \usepackage{textcomp} 
\else 
  \usepackage{unicode-math} 
  \defaultfontfeatures{Scale=MatchLowercase}
  \defaultfontfeatures[\rmfamily]{Ligatures=TeX,Scale=1}
\fi
\usepackage[]{mathpazo}
\ifPDFTeX\else
\fi
\IfFileExists{upquote.sty}{\usepackage{upquote}}{}
\IfFileExists{microtype.sty}{
  \usepackage[]{microtype}
  \UseMicrotypeSet[protrusion]{basicmath} 
}{}
\makeatletter
\@ifundefined{KOMAClassName}{
  \IfFileExists{parskip.sty}{%
    \usepackage{parskip}
  }{
    \setlength{\parindent}{0pt}
    \setlength{\parskip}{6pt plus 2pt minus 1pt}}
}{
  \KOMAoptions{parskip=half}}
\makeatother
\usepackage{xcolor}
\usepackage{listings}
\newcommand{\passthrough}[1]{#1}
\providecommand{\tightlist}{%
  \setlength{\itemsep}{0pt}\setlength{\parskip}{0pt}}
\usepackage[T1]{fontenc}
\usepackage{mathpazo}              
\usepackage[scaled=0.88]{helvet}  
\usepackage{microtype}

\usepackage[letterpaper,margin=1in,headheight=14pt]{geometry}

\usepackage{xcolor}
\definecolor{linknavy}{HTML}{1A3D6D}
\definecolor{rulegray}{HTML}{888888}
\definecolor{codebg}{HTML}{F6F6F4}
\definecolor{codeframe}{HTML}{D9D9D4}
\definecolor{codecmt}{HTML}{6A737D}

\usepackage{titlesec}
\titleformat{\section}{\normalfont\large\bfseries}{\thesection}{0.6em}{}
\titleformat{\subsection}{\normalfont\normalsize\bfseries}{\thesubsection}{0.6em}{}
\titlespacing*{\section}{0pt}{1.4ex plus 1ex minus .2ex}{0.8ex plus .2ex}
\titlespacing*{\subsection}{0pt}{1.1ex plus 1ex minus .2ex}{0.6ex plus .2ex}

\usepackage{enumitem}
\setlist{leftmargin=1.4em,itemsep=0.2ex,topsep=0.4ex}

\usepackage{graphicx}
\usepackage{float}
\usepackage[font=small,labelfont=bf,skip=6pt]{caption}

\usepackage{listings}
\usepackage{hyperref}
\hypersetup{
  colorlinks=true,
  linkcolor=linknavy,
  citecolor=linknavy,
  urlcolor=linknavy,
  breaklinks=true,
  pdfauthor={Jakob Salfeld-Nebgen},
  pdftitle={Governing Actions, Not Agents: Institutional Attestation as a Governance Model for Autonomous AI Systems},
  pdfkeywords={AI agent governance; institutional attestation; zero-trust architecture; policy decision point; cryptographic attestation; verifiable computation; tamper-evident audit},
}
\usepackage{url}

\usepackage{titling}
\pretitle{\begin{center}\LARGE\bfseries}
\posttitle{\par\end{center}\vskip 0.4em}
\preauthor{\begin{center}\large}
\postauthor{\par\end{center}}
\predate{\begin{center}\normalsize\itshape}
\postdate{\par\end{center}}
\usepackage{fancyhdr}
\usepackage{etoolbox}
\renewenvironment{abstract}{%
  \vspace{0.4em}%
  \begin{center}\bfseries\normalsize Abstract\end{center}%
  \begin{quote}\small\noindent\ignorespaces}{%
  \end{quote}\vspace{0.6em}}
\ifLuaTeX
  \usepackage{selnolig}  
\fi
\IfFileExists{bookmark.sty}{\usepackage{bookmark}}{\usepackage{hyperref}}
\IfFileExists{xurl.sty}{\usepackage{xurl}}{} 
\hypersetup{
  pdftitle={Governing Actions, Not Agents: Institutional Attestation as a Governance Model for Autonomous AI Systems},
  hidelinks,
  pdfcreator={LaTeX via pandoc}}

\title{Governing Actions, Not Agents: Institutional Attestation as a
Governance Model for Autonomous AI Systems}
\author{Jakob Salfeld-Nebgen\\
\texttt{metaphora.ai}}
\date{Proof-of-concept implementation:
\url{https://github.com/jsalfeld/zta-hub}}

\makeatletter
\renewenvironment{thebibliography}[1]
  {\subsection*{\refname}\@mkboth{}{}%
   \list{\@biblabel{\@arabic\c@enumiv}}%
        {\settowidth\labelwidth{\@biblabel{#1}}%
         \leftmargin\labelwidth \advance\leftmargin\labelsep
         \usecounter{enumiv}\let\p@enumiv\@empty
         \renewcommand\theenumiv{\@arabic\c@enumiv}}%
   \sloppy\clubpenalty4000 \@clubpenalty\clubpenalty
   \widowpenalty4000 \sfcode`\.\@m}%
  {\def\@noitemerr{\@latex@warning{Empty `thebibliography' environment}}\endlist}
\makeatother

\begin{document}
\maketitle
\begin{abstract}
Autonomous AI agents may begin to perform consequential, irreversible
actions such as clinical prescribing and production software deployment.
This paper observes that human institutions have governed powerful
autonomous actors not by monitoring their reasoning but by requiring
independently attested evidence at the point of consequential action. We
formalise this institutional pattern as a computational governance model
for AI agent systems. Under the proposed model, an agent retains full
autonomy over planning and reasoning but holds no execution authority
over designated high-risk actions. Execution is conditional on
preconditions that are each independently attested by a separate
authoritative source, cryptographically bound to a declared intent, and
evaluated by a deterministic policy. Decisions are recorded in a
tamper-evident log amenable to independent re-verification. We present a
proof-of-concept implementation and illustrate the model with examples from
software deployment and clinical prescribing.

\smallskip\noindent\textbf{Keywords:} AI agent governance; institutional
attestation; zero-trust architecture; policy decision point;
cryptographic attestation; verifiable computation; tamper-evident audit.
\end{abstract}

\hypertarget{introduction}{%
\subsection{1. Introduction}\label{introduction}}

Large language model agents now plan multi-step tasks, invoke external
tools, and produce side effects in systems of record without per-step
human approval. This raises a governance question: under what conditions
should such a system be permitted to act?

Existing approaches typically instrument the agent's runtime,
intercepting tool calls, classifying behaviour, and enforcing policy
over observed execution context. These mechanisms are effective for
operational constraints --- restricting which tools may be called,
enforcing rate limits, blocking known-dangerous parameter patterns.
However, they operate on the mechanics of tool invocation: the tool
name, its parameters, and the shape of its response. For actions whose
correctness depends on facts about the state of the world --- whether a
drug interaction has been checked, whether a build has passed, whether a
licence is valid --- the relevant information resides in authoritative
external systems that the agent's runtime does not consult.

This paper proposes a governance model drawn from an older and
well-tested source: institutional governance of consequential acts.
Human institutions have governed powerful autonomous actors ---
physicians, judges, financial officers --- not by monitoring their
deliberation but by imposing requirements at the point of action. A
controlled-substance prescription requires a verified patient record, a
drug-interaction clearance, and a valid DEA registration, each attested
by an independent authoritative source. The actor's reasoning is not
governed; the act is, through independently attested evidence.

We formalise this pattern as a computational governance model for AI
agent systems. The model rests on three commitments:

\begin{itemize}
\tightlist
\item
  \textbf{Actions are governed, not agents.} Governance is applied at
  the boundary where an agent produces an irreversible side effect, not
  at the boundary of its reasoning or planning.
\item
  \textbf{Agents retain autonomy.} The agent discovers governance
  requirements at run time and assembles the necessary evidence without
  modification to its internal control flow.
\item
  \textbf{Evidence is independently verifiable.} An action is permitted
  only when each of its preconditions has been attested by a distinct,
  independent external authority, bound to the specific intent, and
  evaluated by a deterministic policy. No single party --- neither the
  agent nor any one gatekeeper --- supplies the full basis for a
  decision; the evidence is assembled from separate attesters and the
  resulting decision is re-verifiable by any third party.
\end{itemize}

The primitives underlying this model are individually well-established,
and several concurrent 2026 preprints develop closely related ideas:
removing execution authority from the agent and binding an action to a
verified intent behind a brokered admission boundary \cite{seb,sab};
deterministic pre-action authorisation with a tamper-evident audit
record \cite{oap}; and an institutional, separation-of-powers framing of
agent governance \cite{logic-monopoly}. A related line of work examines
intent-to-execution integrity \cite{intent-execution}. Our aim is to articulate their
composition into a single institutional governance model for the
agent-action boundary, grounded in long-standing institutional precedent
and illustrated with a proof-of-concept implementation that has been publicly available since May 2026 and was developed independently of these concurrent efforts.

\hypertarget{institutional-governance-as-a-model}{%
\subsection{2. Institutional Governance as a
Model}\label{institutional-governance-as-a-model}}

The institutional pattern exhibits several properties that serve as
design requirements for the computational model.

\textbf{Independent attestation.} No single party's attestation is
sufficient. Each precondition is verified by an independent authority
--- a licensing body, a records system, a testing service --- with its
own basis for judgement and its own credentials. The decision-maker
evaluates the collected attestations but does not produce them. Trust is
distributed across independent attesters rather than concentrated in the
actor or in a single gatekeeper. This realises the separation-of-duty
principle formalised by Clark and Wilson \cite{clark-wilson}.

\textbf{Transaction binding.} Attestations are bound to a specific act.
A notarised document identifies the date, the parties, and the
transaction. A prior attestation cannot be substituted for the current
one. This prevents both replay and the detachment of evidence from the
context that gave it meaning.

\textbf{Deterministic evaluation against stated rules.} The decision
follows from applying declared rules to attested facts. The rule is
stated in advance, applied uniformly, and its application is
reproducible. Given identical attested facts, the same rule yields the
same decision. This corresponds to the policy decision point
architecture described in XACML \cite{xacml} and may be realised using
deterministic policy languages such as Cedar \cite{cedar} or Rego \cite{rego}.

\textbf{Permanent, independently auditable record.} The decision, the
underlying evidence, and the rule applied are recorded such that any
authorised third party can later inspect and re-verify the decision.
This draws on tamper-evident log constructions \cite{merkle} and transparency
log architectures such as certificate transparency \cite{ct-rfc6962}. Existing
frameworks for supply-chain attestation, such as in-toto \cite{in-toto} and
SCITT \cite{scitt}, address closely related problems in software provenance
and provide relevant precedent for the attestation and transparency
mechanisms described here.

\textbf{Autonomy of the actor.} The actor's deliberation is not
governed. Governance engages only at the point of action, and then only
by requiring evidence. The pattern is compatible with discretion and
competence because it constrains consequences, not reasoning.

These properties are individually well-established in the security
literature: zero-trust verification \cite{nist-zta}, capability security \cite{saltzer-schroeder,miller}, Byzantine fault tolerance \cite{lamport-byzantine}, and the reference monitor
concept \cite{anderson}. Our aim here is to compose them into a unified
governance model, motivated by institutional precedent and applied to
the agent-action boundary.

\hypertarget{the-computational-model}{%
\subsection{3. The Computational Model}\label{the-computational-model}}

We formalise the institutional pattern as a governance architecture for
AI agent systems. A proof-of-concept implementation, the Zero-Trust Action
Hub
(\href{https://github.com/jsalfeld/zta-hub}{github.com/jsalfeld/zta-hub}),
illustrates this model; we describe the abstract mechanisms and refer to
the implementation for engineering detail.

\hypertarget{the-courier-pattern}{%
\subsubsection{3.1 The Courier Pattern}\label{the-courier-pattern}}

In conventional tool-calling architectures, the agent holds credentials
and executes actions directly. Under the model proposed here, the agent
holds no execution authority over governed actions. Instead, it operates
as a courier through the following steps:

\begin{enumerate}
\def\labelenumi{\arabic{enumi}.}
\tightlist
\item
  \textbf{Intent declaration.} The agent requests authorisation for a
  specific governed action. The governance hub issues a unique,
  cryptographically random intent identifier --- a binding token to
  which all subsequent attestations must refer --- and returns the list
  of required attestations.
\item
  \textbf{Evidence collection.} The agent contacts the required
  authoritative sources --- independent services termed oracles --- and
  collects signed attestations that each precondition holds. Each oracle
  verifies one condition and returns the result signed with its own
  private key, bound to the intent identifier.
\item
  \textbf{Submission and evaluation.} The agent presents the collected
  attestations to the governance hub. The hub verifies each signature
  against pre-registered public keys, confirms intent binding, and
  evaluates a deterministic policy over the attested facts. The default
  posture is deny.
\item
  \textbf{Conditional authorisation.} If the policy permits, the hub
  either executes the action on the agent's behalf or issues a signed,
  narrowly-scoped capability token. The decision is appended to a
  tamper-evident audit log.
\end{enumerate}

The agent assembles evidence but cannot fabricate it, as it does not
hold the oracles' signing keys. Authorisation derives from verified
evidence evaluated against policy, not from the agent's standing
privilege.

\hypertarget{multi-party-attestation-and-intent-binding}{%
\subsubsection{3.2 Multi-Party Attestation and Intent
Binding}\label{multi-party-attestation-and-intent-binding}}

Each precondition is checked by an independent oracle with its own
asymmetric key pair (e.g., Ed25519 \cite{ed25519}). No party, including the
governance hub, can produce another party's attestation. Attestation
envelopes may follow established formats such as DSSE \cite{dsse} to
facilitate interoperability. Every attestation includes the intent
identifier issued at declaration, and the oracle signs the intent
identifier together with its source identifier, expiry, and payload as a
single canonical envelope. The identifier serves as a cryptographic
binding between the specific action request and the evidence collected
for it; because it falls within the signed envelope, neither the binding
nor the attested facts can be altered without invalidating the
signature, and an attestation produced for one intent is rejected if
presented for another. This prevents both the reuse of evidence across
actions and the substitution of attestations between unrelated requests.
Each attestation additionally carries an expiry, also within the signed
envelope; the hub rejects any attestation presented after its expiry.
This bounds the interval between the moment a precondition is checked
and the moment the action executes, ensuring that evidence is freshly
collected for each action rather than carried over from an earlier
context.

\hypertarget{deterministic-policy-and-verified-computation}{%
\subsubsection{3.3 Deterministic Policy and Verified
Computation}\label{deterministic-policy-and-verified-computation}}

Attested facts are assembled into a policy context and evaluated against
a declarative policy. Because evaluation is deterministic, the decision
is reproducible from the recorded inputs.

When a precondition involves a computation rather than an external
lookup --- for example, a dosage calculation or an inference over held
data --- the agent may execute approved code and submit a proof of
correct execution (via trusted-execution-environment attestation or a
zero-knowledge argument \cite{gmr}) against a registered, audited code
hash. The hub verifies the proof and admits only the verified output
into the policy context.

\hypertarget{tamper-evident-audit-and-action-composition}{%
\subsubsection{3.4 Tamper-Evident Audit and Action
Composition}\label{tamper-evident-audit-and-action-composition}}

Each decision is appended to a tamper-evident audit log constructed
using hash-chain or Merkle-tree techniques \cite{merkle}. Every entry includes
the intent identifier, the action type, the signed receipt, and a
cryptographic commitment to prior entries. Altering a past record
invalidates subsequent entries. Transparency log services such as those
used in certificate transparency \cite{ct-rfc6962} or supply-chain integrity
frameworks \cite{in-toto,scitt} provide established infrastructure for this
purpose. A third party can re-verify any decision by checking oracle
signatures, confirming intent binding, re-evaluating the policy, and
walking the log.

A successful action yields a signed execution receipt that may serve as
a precondition for a subsequent action. This permits governance to
compose: the policy for a later action can require verifiable evidence
that a prior action was itself governed and executed.

\hypertarget{governance-discovery}{%
\subsubsection{3.5 Governance Discovery}\label{governance-discovery}}

Requirements for each governed action are published as machine-readable
skill contracts specifying risk classification, required attestations,
oracle endpoints, and input/output schemas. The agent reads these at run
time and assembles evidence accordingly. Adding or modifying a governed
action requires no change to the agent. This makes the governance
surface scalable along the dimensions that matter organisationally: new
governed actions, new attesters, and additional oracles can be
introduced without modifying the agent or the hub's evaluation logic,
and governance composes across actions through execution receipts
(\S3.4). Because attesters are independent, trust is distributed across
organisational and jurisdictional boundaries rather than concentrated in
one authority. Scalability here is in governance coverage, not in
per-action throughput; the cost of collecting multiple attestations per
action bounds applicability to high-volume routine operations.

\hypertarget{examples}{%
\subsection{4. Examples}\label{examples}}

\hypertarget{software-deployment}{%
\subsubsection{4.1 Software Deployment}\label{software-deployment}}

Consider an AI agent that has completed a feature and seeks to deploy it
to a production environment. Under this model,
\passthrough{\lstinline!deploy\_to\_production!} is a governed action
requiring three independently attested preconditions.

\begin{figure}[htbp]
\centering
\includegraphics[width=\linewidth]{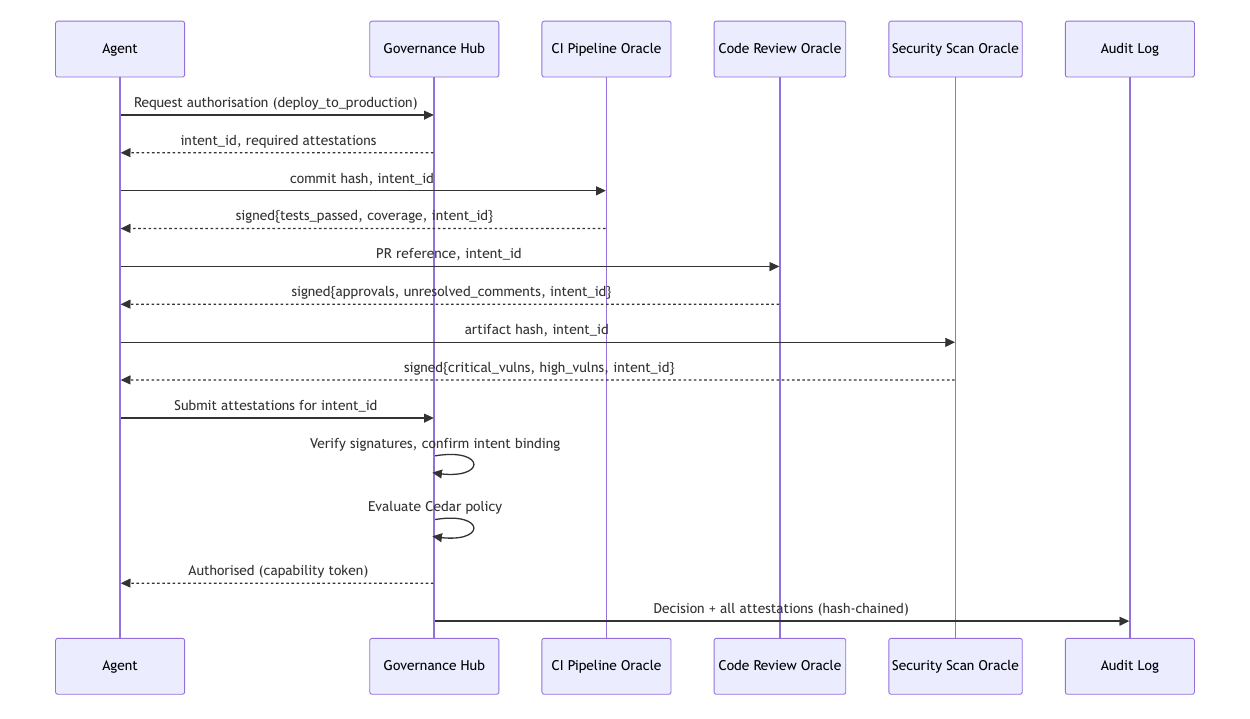}
\caption{Governed \texttt{deploy\_to\_production}: the agent collects independently signed attestations from the CI, code-review, and security-scan oracles, each bound to the issued intent identifier, before the hub verifies and authorises.}
\label{fig:deploy}
\end{figure}

Each oracle returns a signed attestation envelope containing its
verified facts and the intent identifier. For example, the CI oracle
returns:

\begin{lstlisting}
{
  "source_id": "ci_pipeline",
  "intent_id": "int-7f3a",
  "expires_at": "2026-06-23T12:05:00Z",
  "payload": {"tests_passed": true, "coverage": 94.2},
  "signature": "<Ed25519 signature over {source_id, intent_id, expires_at, payload}>"
}
\end{lstlisting}

Upon receiving the attestations, the hub executes the following
verification pipeline before policy evaluation:

\begin{enumerate}
\def\labelenumi{\arabic{enumi}.}
\tightlist
\item
  \textbf{Signature verification.} For each attestation, the hub
  retrieves the oracle's pre-registered public key by
  \passthrough{\lstinline!source\_id!} and verifies the Ed25519
  signature over the canonical envelope --- the source identifier,
  intent identifier, expiry, and payload. An invalid or unrecognised
  signature rejects the request.
\item
  \textbf{Intent binding.} The hub confirms that the
  \passthrough{\lstinline!intent\_id!} in each attestation matches the
  intent declared at step 1. A mismatch --- indicating a replayed or
  substituted attestation --- rejects the request.
\item
  \textbf{Freshness.} The hub confirms that each attestation is
  presented before its expiry. An expired attestation rejects the
  request, so stale evidence cannot be carried over from an earlier
  action.
\item
  \textbf{Completeness.} The hub confirms that all attestations required
  by the skill contract are present. A missing attestation rejects the
  request.
\item
  \textbf{Context assembly.} The verified payload fields from all
  attestations are merged into a single policy context; where two
  oracles would contribute the same field name, the field is qualified
  by its source identifier. Only data extracted from signature-verified,
  intent-bound attestations enters this context.
\end{enumerate}

The hub then evaluates the Cedar policy over the assembled context:

\begin{lstlisting}
permit(
    principal,
    action == Action::"deploy_to_production",
    resource
) when {
    context.tests_passed == true &&
    context.coverage >= 80 &&
    context.approvals >= 2 &&
    context.unresolved_comments == 0 &&
    context.critical_vulns == 0 &&
    context.high_vulns == 0
};
\end{lstlisting}

The decision and all supporting evidence --- including the original
signed attestations --- are recorded in the audit log and remain
independently re-verifiable.

\hypertarget{clinical-prescribing}{%
\subsubsection{4.2 Clinical Prescribing}\label{clinical-prescribing}}

Consider an AI clinical agent that determines a patient requires a
controlled substance. Under this model,
\passthrough{\lstinline!prescribe\_medication!} is a governed action
requiring three independently attested preconditions.

\begin{figure}[htbp]
\centering
\includegraphics[width=\linewidth]{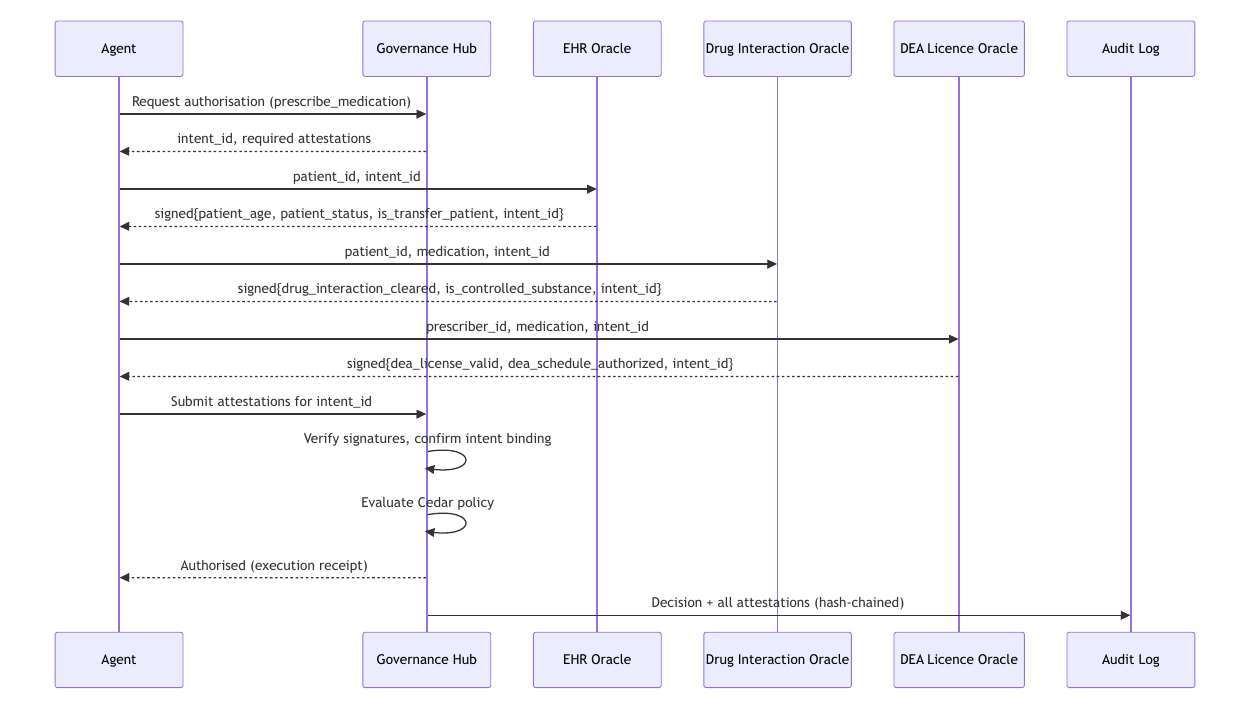}
\caption{Governed \texttt{prescribe\_medication}: the agent collects independently signed attestations from the EHR, drug-interaction, and DEA-licence oracles, each bound to the issued intent identifier, before the hub verifies and authorises.}
\label{fig:prescribe}
\end{figure}

Each oracle returns a signed attestation envelope. For example, the drug
interaction oracle returns:

\begin{lstlisting}
{
  "source_id": "drug_interaction_oracle",
  "intent_id": "int-9d2e",
  "expires_at": "2026-06-23T12:05:00Z",
  "payload": {"drug_interaction_cleared": true, "is_controlled_substance": true},
  "signature": "<Ed25519 signature over {source_id, intent_id, expires_at, payload}>"
}
\end{lstlisting}

The hub executes the same verification pipeline as in Section 4.1:
signature verification against pre-registered public keys, intent
binding confirmation, completeness check, and context assembly from
verified payloads only. It then evaluates:

\begin{lstlisting}
permit(
    principal,
    action == Action::"prescribe_medication",
    resource
) when {
    context.drug_interaction_cleared == true &&
    context.dea_license_valid == true &&
    context.patient_age >= 18 &&
    (if context.is_controlled_substance == true then
        context.dea_schedule_authorized == true else true) &&
    (if context.is_transfer_patient == true then
        context.prior_actions.release_medical_records.status == "executed"
        else true)
};
\end{lstlisting}

The final clause illustrates action composition. If the patient is a
transfer patient, the policy requires that a prior governed action,
\passthrough{\lstinline!release\_medical\_records!}, was executed under
the same governance regime. The evidence for this is a signed execution
receipt issued by the hub upon completion of the prior action --- itself
a verified attestation that enters the policy context through the same
signature-verification pipeline, surfaced under
\passthrough{\lstinline!prior\_actions.release\_medical\_records!}.

\hypertarget{limitations-and-scope}{%
\subsection{5. Limitations and Scope}\label{limitations-and-scope}}

\textbf{Coverage depends on action classification.} Only designated
actions are governed. The model does not determine which actions are
high-risk; that classification is an organisational judgement and a
prerequisite for deployment.

\textbf{Oracle integrity is a foundational assumption.} The hub verifies
that an oracle signed an attestation, not that the attested fact is true
of the world. A compromised oracle or a leaked signing key undermines
the guarantees for the conditions it covers. Oracle independence,
operational integrity, and key custody are load-bearing assumptions.

\textbf{Time-of-check to time-of-use.} A precondition is attested at one
moment and the action executes at another; in the interval, the
underlying fact may change --- a licence may be revoked, a build
superseded, a drug interaction newly identified. The expiry carried by
each attestation controls this risk by bounding the interval: evidence
expires and must be re-collected, so the gap between check and use is
limited to the validity window rather than left open indefinitely. The
risk is bounded, not eliminated. Within the window the world may still
change, and for facts that can change abruptly the window must be set
short, or supplemented by a revocation check or re-attestation
immediately before execution.

\textbf{Intent-matching is not assured.} The model verifies that
declared preconditions for a declared action are satisfied. It does not
verify that the agent's declared intent corresponds to the human
principal's actual goal. An agent that requests a correctly governed
action for the wrong reason --- or that decomposes a harmful goal into
individually legitimate governed steps --- is not prevented by this
model. This residual risk is shared with institutional governance: a
physician who orders a legitimate prescription for an illegitimate
purpose satisfies the procedural requirements while violating their
spirit. Complementary measures, including human-in-the-loop review for
novel or unusual patterns, compose with evidence-gating to address this
gap.

\textbf{Policy correctness is not assured by the model.} The model
guarantees faithful enforcement of a stated policy. Whether the policy
expresses the correct rule is an organisational responsibility.

\textbf{Governance of the path versus governance of the boundary.} The
model governs whether preconditions hold at the point of action; it does
not prescribe the sequence of steps the agent takes to reach that point.
When the ordering or manner of steps is itself safety-relevant --- for
instance, when a specific protocol must be followed regardless of
outcome --- prescriptive workflow governance is more appropriate. The
two approaches address different concerns and may be combined.

\textbf{Operational cost.} Collecting attestations from multiple
external services introduces latency and engineering overhead. This cost
is proportionate for high-stakes actions but may be disproportionate for
routine operations.

\textbf{Relevance to identified risks.} The model is particularly
relevant to the rogue agent problem \cite{owasp-agentic}, where a compromised agent
operates within its authorised scope while pursuing adversarial
objectives --- a scenario in which runtime behavioural monitoring is
structurally ineffective because individual actions conform to expected
patterns. The model also reaches a failure mode that is structurally
inaccessible to the prevailing safety layers. When an action is locally
well-formed --- the right tool, plausible parameters --- but its
correctness depends on a fact about the world (a build passed, a licence
is valid, an interaction was checked), alignment, runtime behavioural
monitoring, and sandboxing cannot adjudicate it, because that fact is
absent from the execution context they observe. Evidence-gating admits
the action only against the fact itself, independently attested and
beyond the agent's control to fabricate. This control extends only to
governed actions and does not, on its own, address an agent pursuing an
adversarial objective through individually legitimate governed steps. It
also provides a compliance path for audit requirements that demand
independently verifiable decision records \cite{eu-ai-act}, as the signed
attestations and tamper-evident log support third-party reconstruction
without reliance on operator attestation.

\textbf{Relationship to runtime governance.} Runtime governance
mechanisms --- tool-call interception, deny-lists, rate limiting,
response scanning --- enforce operational constraints across the broad
surface of agent activity. They are effective for rules expressible in
terms of tool-call mechanics: which tools may be invoked, how
frequently, and with what parameter shapes. The model proposed here
addresses a distinct requirement: establishing whether the substantive,
real-world preconditions for a consequential action have been
independently verified. The relevant facts --- whether a build passed,
whether a licence is valid, whether a drug interaction was checked ---
reside in authoritative external systems and are not available in the
tool-call execution context. The two approaches operate at different
levels of abstraction. A deployment that combines runtime governance for
operational breadth with institutional attestation for high-stakes
actions would address both concerns.

\hypertarget{conclusion}{%
\subsection{6. Conclusion}\label{conclusion}}

For consequential actions, the governance question is not whether an
agent may invoke a particular tool, but whether the real-world
preconditions for the action have been independently established. Human
institutions have addressed this question by requiring independently
attested evidence at the point of action rather than by governing the
actor's reasoning. This paper formalises that institutional pattern as a
computational governance model: intent declaration, multi-party
cryptographic attestation, transaction binding, deterministic policy
evaluation, and tamper-evident audit. The constituent primitives are
well-established, and several concurrent efforts develop related
mechanisms; the aim of this paper is to compose them into a coherent
institutional governance model for autonomous AI systems, grounded in
institutional precedent.

The open problem is not primarily technical. It is the question of which
actions warrant this level of governance and what evidence should be
required --- a question that remains one of judgement and
accountability.


\begin{thebibliography}{99}

\bibitem{clark-wilson} D. D. Clark and D. R. Wilson, ``A Comparison of Commercial and Military Computer Security Policies,'' IEEE Symposium on Security and Privacy, 1987.

\bibitem{xacml} OASIS, ``eXtensible Access Control Markup Language (XACML) Version 3.0,'' OASIS Standard, 2013.

\bibitem{cedar} J. Cutler et al., ``Cedar: A New Language for Expressive, Fast, Safe, and Analyzable Authorization,'' Proc. ACM on Programming Languages (OOPSLA), 2024.

\bibitem{rego} The Open Policy Agent Authors, ``Rego Policy Language,'' https://www.openpolicyagent.org/docs/latest/policy-language/.

\bibitem{merkle} R. C. Merkle, ``A Digital Signature Based on a Conventional Encryption Function,'' Advances in Cryptology (CRYPTO '87), 1988.

\bibitem{ct-rfc6962} B. Laurie, A. Langley, and E. Kasper, ``Certificate Transparency,'' RFC 6962, Internet Engineering Task Force, 2013.

\bibitem{in-toto} S. Torres-Arias, H. Afzali, T. K. Kuppusamy, R. Curtmola, and J. Cappos, ``in-toto: Providing farm-to-table guarantees for bits and bytes,'' USENIX Security Symposium, 2019.

\bibitem{scitt} IETF SCITT Working Group, ``An Architecture for Trustworthy and Transparent Digital Supply Chains,'' Internet-Draft, 2024.

\bibitem{nist-zta} S. Rose, O. Borchert, S. Mitchell, and S. Connelly, ``Zero Trust Architecture,'' NIST Special Publication 800-207, 2020.

\bibitem{saltzer-schroeder} J. H. Saltzer and M. D. Schroeder, ``The Protection of Information in Computer Systems,'' Proceedings of the IEEE, vol.~63, no. 9, 1975.

\bibitem{miller} M. S. Miller, ``Robust Composition: Towards a Unified Approach to Access Control and Concurrency Control,'' Ph.D.~dissertation, Johns Hopkins University, 2006.

\bibitem{lamport-byzantine} L. Lamport, R. Shostak, and M. Pease, ``The Byzantine Generals Problem,'' ACM Transactions on Programming Languages and Systems, vol.~4, no. 3, 1982.

\bibitem{anderson} J. P. Anderson, ``Computer Security Technology Planning Study,'' Technical Report ESD-TR-73-51, US Air Force Electronic Systems Division, 1972.

\bibitem{ed25519} D. J. Bernstein, N. Duif, T. Lange, P. Schwabe, and B.-Y. Yang, ``High-Speed High-Security Signatures,'' Journal of Cryptographic Engineering, vol.~2, 2012.

\bibitem{dsse} E. Engelke and S. Torres-Arias, ``Dead Simple Signing Envelope,'' https://github.com/secure-systems-lab/dsse, 2021.

\bibitem{gmr} S. Goldwasser, S. Micali, and C. Rackoff, ``The Knowledge Complexity of Interactive Proof Systems,'' SIAM Journal on Computing, vol.~18, no. 1, 1989.

\bibitem{owasp-agentic} OWASP, ``Top 10 for Agentic Applications,'' OWASP GenAI Security Project, 2025. See ASI10: Rogue Agents.

\bibitem{eu-ai-act} European Parliament and Council, ``Regulation (EU) 2024/1689 laying down harmonised rules on artificial intelligence (Artificial Intelligence Act),'' Article 12: Record-keeping, 2024.

\bibitem{oap} U. Uchibeke, ``Before the Tool Call: Deterministic Pre-Action Authorization for Autonomous AI Agents,'' arXiv preprint arXiv:2603.20953, 2026.

\bibitem{seb} J. He and D. Yu, ``Sovereign Execution Broker: Enforcing Certificate-Bound Authority in Agentic Control Planes,'' arXiv preprint arXiv:2606.20520, 2026.

\bibitem{logic-monopoly} A. Ruan, ``From Logic Monopoly to Social Contract: Separation of Power and the Institutional Foundations for Autonomous Agent Economies,'' arXiv preprint arXiv:2603.25100, 2026.

\bibitem{intent-execution} W. Qu, M. Xu, P. Wang, S. Zhai, J. Zhang, and D. Song, ``Securing LLM Agents Need Intent-to-Execution Integrity,'' arXiv preprint arXiv:2605.16976, 2026.

\bibitem{sab} J. He and D. Yu, ``Sovereign Assurance Boundary: Certificate-Bound Admission for Agentic Infrastructure,'' arXiv preprint arXiv:2606.11632, 2026.

\end{thebibliography}
\end{document}